\documentclass[11pt,a4paper]{article}
\usepackage[utf8]{inputenc} %
\usepackage{times,latexsym}
\usepackage[hyphens]{url}
\usepackage{breakurl}
\usepackage[T1]{fontenc}

\usepackage[acceptedWithA]{tacl2021v1} %

\usepackage{xspace,mfirstuc,tabulary}

\newif\iftaclinstructions
\taclinstructionsfalse %
\iftaclinstructions
\renewcommand{\confidential}{}
\renewcommand{\anonsubtext}{(No author info supplied here, for consistency with
TACL-submission anonymization requirements)}
\newcommand{\instr}
\fi

\iftaclpubformat %

\else

\fi

\usepackage{graphicx}

\usepackage{color, soul}
\usepackage{booktabs}
\usepackage{amssymb}
\usepackage{enumitem}
\usepackage[linguistics]{forest}

\hypersetup{
   breaklinks=true,   %
   colorlinks=true,   %
}

\newcommand{\aam}{$_{\mbox{A}}$\xspace}
\newcommand{\avm}{$_{\mbox{V}}$\xspace}

\newcommand{\valla}{{\sc Valla}\xspace}

\newcommand{\RNum}[1]{\uppercase\expandafter{\romannumeral #1\relax}}

\newcommand{\ngram}{Ngram\xspace}
\newcommand{\ngrama}{Ngram\aam}
\newcommand{\ngramv}{Ngram\avm}

\graphicspath{ {./images/} }

\title{On the State of the Art in Authorship Attribution and\\Authorship Verification}

\author{
 Jacob Tyo$^{1,2}$, Bhuwan Dhingra$^3$, and Zachary C. Lipton$^2$ \\
 $^1$DEVCOM Army Research Laboratory \\
 \texttt{jacob.p.tyo.civ@army.mil} \\
 $^2$Carnegie Mellon University \\
 \texttt{zlipton@cmu.edu} \\
 $^3$Duke University \\
 \texttt{bdhingra@cs.duke.edu} \\
}

\begin{document}
\maketitle

\begin{abstract}
    
Despite decades of research
on authorship attribution (AA) 
and authorship verification (AV),
inconsistent dataset splits/filtering 
and mismatched evaluation methods 
make it difficult to assess the state of the art.
In this paper, 
we present a survey of the fields, 
resolve points of confusion,
introduce \valla that 
standardizes and benchmarks AA/AV datasets and metrics, 
provide a large-scale empirical evaluation,
and provide apples-to-apples comparisons 
between existing methods.
We evaluate eight promising methods on fifteen datasets
(including distribution shifted challenge sets) 
and introduce a new large-scale dataset 
based on texts archived by Project Gutenberg. 
Surprisingly, we find that a traditional \ngram-based model 
performs best on 5 (of 7) AA tasks,
achieving an average macro-accuracy of $76.50\%$
(compared to $66.71\%$ for a BERT-based model).
However, on the two AA datasets with the greatest number of words per author, 
as well as on the AV datasets, 
BERT-based models perform best. 
While AV methods are easily applied to AA,
they are seldom included as baselines in AA papers. 
We show that through the application of hard-negative mining, 
AV methods are competitive alternatives to AA methods.
\valla and all experiment code can be found here:
\url{https://github.com/JacobTyo/Valla}

\end{abstract}

\section{Introduction}
\label{sec:intro}

The statistical analysis of variations in literary style 
between one writer or genre and another,
commonly known as \emph{stylometry},
dates back as far as 500 AD, 
when the Hebrew Old Testament was studied 
and standardized by Tiberias at Palestine~\citep{binongo1996stylometry}. 
Computer-assisted stylometry first emerged in the early 1960s, 
when \citet{mosteller1963inference} explored the foundations 
of computer-assisted authorship analysis.
Today automated tools for authorship analysis are common,
finding practical use in the justice system
to analyze evidence~\citep{koppel2008authorship},  
among social media companies 
to detect compromised accounts~\citep{barbon2017authorship},
to link online accounts that belong to the same individual~\citep{sinnott2021linking},
and in a variety of contexts 
to detect plagiarism~\citep{stamatatos2011plagiarism}.

In the modern Natural Language Processing (NLP) literature,
two primary problem formulations dominate the empirical study
of methods for determining the authorship 
of anonymous or disputed texts: 
Authorship Attribution (AA) and Authorship Verification (AV). 
In AA, the learner is given representative texts
for a canonical set of authors in advance,
and expected to attribute a new previously unseen text
of unknown authorship to one of these a priori known authors. 
In AV, the learner faces a more general problem:
given two texts, predict whether they were written 
by the same author or not.

While both problems have received considerable attention
\citep{murauer2021developing, altakrori2021topic, kestemont2021overview},
the state of the art is difficult to assess
owing to inconsistencies in the datasets, splits,
performance metrics,  
variations in the framing of domain shift across studies,
and the lack of large-scale datasets. 
For example, a recent survey paper~\citep{neal2017surveying} indicates 
that the state-of-the-art method is based 
on the Partial Matching (PPM) text compression scheme
and the cross-entropy of each text 
with respect to the PPM categories. 
By contrast, the PAN-2021 competition~\citep{kestemont2021overview} 
indicates that the state of the art is a hierarchical bi-directional LSTM 
with learned-CNN text encodings. 
Recent work~\citep{fabien2020bertaa} concludes 
that the transformer-based language model BERT 
is the highest performing AA method. 
A recent analysis paper~\citep{altakrori2021topic} argue 
that the traditional approach of character n-grams and masking
remains the best methodology to this day.
Each of these sources compares methods against different baselines, 
on different datasets (sometimes on just a single small dataset), 
and with different problem variations 
(such as same-topic, cross-topic, cross-genre, cross-language, etc.).

In this paper, we start by sorting out this fragmented prior work
through a brief survey of the literature.
Then, to present a unified evaluation,
we introduce The \valla Benchmark. %
This benchmark provides standardized versions 
of all the common AA and AV datasets with uniform evaluation metrics and
standardized domain shifted test sets.
Additionally, we introduce a new large-scale dataset
based on public domain books sourced from Project Gutenberg 
for both tasks.
Then using this benchmark,
we present an extensive evaluation 
of eight common AA and AV methods on their respective
datasets with and without
domain shift.
We also make comparisons \emph{between} AA and AV methods
where applicable.

Although much recent work indicates 
that AA and AV are not benefiting 
from the recent advancements 
of pre-trained language models (such as BERT and GPT-2)
commensurate with the gains seen elsewhere in NLP
\citep{kestemont2021overview, altakrori2021topic, murauer2021developing, tyo2021siamese, peng2021encoding, futrzynski2021author},
we show that this narrative only appears to apply to datasets with a limited number of words per class.
Furthermore, BERT-based models achieve new state-of-the-art macro-accuracy 
on the IMDb62 ($98.80\%$) and Blogs50 ($74.95\%$) datasets
and set the benchmark on our newly introduced Gutenberg dataset.

Although the applicability of AV methods to AA problems is frequently mentioned, 
few studies place these methods in competition. 
We provide this comparison and find that, on AA problems, AA methods outperform AV methods by over %
$15\%$ macro-accuracy on average ($76.80\%$ and $59.89\%$ average macro-accuracy for BERT\aam and BERT\avm respectively across five datasets). 
However, hard negative mining improves the performance of AV models in the AA setting, increasing the AA performance of BERT\avm from $67.21\%$ to $72.42\%$ macro-accuracy on the tested dataset, making it a competitive alternative. 

In summary, we contribute the following:
\setlist{nolistsep}
\begin{itemize}[noitemsep]
    \item A survey of AA and AV.  %
    \item An argument for adopting 
    macro-averaged accuracy and AUC as the key metrics for AA and AV, respectively.
    \item A benchmark that standardizes AA and AV datasets, and places them in competition.
    \item State-of-the-art accuracy on the IMDb62 ($98.81\%$) and Blogs50 ($74.95\%$) datasets. 
    \item A new large-scale dataset with long average text length.
    \item An evaluation of eight high-performing AA and AV methods
    on fifteen datasets, assessing the state of the art for both problems.
    \item Evidence of the efficacy of hard-negative mining 
    when applying AV methods to AA problems.
\end{itemize}

\begin{table*}[t]
\centering
\small
\begin{tabular}{@{}ccccccccccccc@{}}
\toprule
Dataset & \begin{tabular}[c]{@{}c@{}}Text\\ Type\end{tabular} & \begin{tabular}[c]{@{}c@{}}Typical\\ Setting\end{tabular} & iid & $\times_t$ & $\times_g$ & $\times_a$ & $D$ & $A$ & $W$ & $D/A$ & $W/D$ & imb \\ \midrule
CCAT50 & News & AA & $\checkmark$ & --- & --- & --- & 5k & 50 & 2.5M & 100 & 506 & 0 \\
CMCC & Various & AA & $\checkmark$ & $\checkmark$ & $\checkmark$ & --- & 756 & 21 & 454k & 36 & 601 & 0 \\
Guardian & Opinion & AA & $\checkmark$ & $\checkmark$ & $\checkmark$ & --- & 444 & 13 & 467k & 34 & 1052 & 6.7 \\
IMDb62 & Reviews & AA & $\checkmark$ & --- & --- & --- & 62k & 62 & 21.6M & 1000 & 349 & 2.6 \\
Blogs50 & Blogs & AA & $\checkmark$ & --- & --- & --- & 66k & 50 & 8.1M & 1324 & 122 & 553 \\
BlogsAll & Blogs & AV & $\checkmark$ & --- & --- & --- & 520k & 14k & 121.6M & 37 & 233 & 90 \\
PAN20 \& 21 & Various & AV & $\checkmark$ & $\checkmark$ & --- & $\checkmark$ & 443k & 278k & 1.7B & 1.6 & 3922 & 2.3 \\
Amazon & Reviews & AV & $\checkmark$ & $\checkmark$ & --- & --- & 1.46M & 146k & 91.9M & 10 & 63 & 0 \\
Gutenberg & Books & AA & $\checkmark$ & --- & --- & $\checkmark$ & 29k & 4.5k & 1.9B & 6 & 66350 & 10.5 \\ \bottomrule
\end{tabular}
\caption{
An overview of datasets used for Authorship Attribution (AA) and Authorship Verification (AV).
iid is an i.i.d. split, $\times_t$ is a cross-topic split, $\times_g$ is a cross-genre split, $\times_a$ is an unknown author split, $D$ is the number of documents, $A$ is the number of authors, $W$ is the number of words, $W/D$ is the average length of documents, $D/A$ is the average number of documents per author, $W/D$ is the average number of words per document, and imb is the imbalance of the dataset measured by the standard deviation of the number of documents per author. $\checkmark$ indicates where the necessary data is available to create a standardized split, whereas --- indicates that the necessary data is not present. 
}
\label{tab:DatasetStatistics}
\end{table*}

\section{Brief Survey of the Literature}
\label{sec:priorWorks}

\citet{neal2017surveying} provide an overview of AA dataset characteristics and traditional AA methods. 
The authors enumerate the wide array of textual features used for AA 
(excluding modern methods such as word embeddings and byte-pair encoding), categorize methods into machine learning, similarity, and probabilistic models, and provide an evaluation of these techniques on a single, small dataset.
They conclude that the predictions using partial matching (PPM) method is the state of the art.
\citet{Bouanani2014authorship} provide a similar survey focusing on the enumeration of hand-engineered features used in AA.
\citet{stamatatos2009survey} discuss traditional AA methods from an instance-based 
(comparing one text to another)
vs a profile-based 
(comparing one text to a representation of all author texts) methodology, including a computational requirement analysis. 

Among notable surveys,
\citet{mekala2018survey} compare the benefits 
of the different traditional textual features;
\citet{argamon2018computational} detail the problems 
with applying many traditional AA methods in real-world scenarios;
\citet{alhijawi2018text} provide a meta-analysis of the field;
and \citet{ma2020towards} point out the lack of advances 
from using transformer-based language models in AA,
calling for more exploration in this area.
Critically, all of these prior surveys 
exclude recent advances due to deep learning, 
such as recurrent neural networks, transformers, 
word embeddings, and byte-pair encoding. 
In this brief survey, 
we briefly cover more traditional techniques,
and then discuss recent deep-learning-based approaches.

So far, we have outlined the work on AA surveys,
but there are none to be found that focus on AV. 
The PAN competition overview~\cite{kestemont2021overview}
is the closest thing that can be viewed as a survey on AV,
but it is limited only to what was seen in the competitions. 
Notably, each year's competition focuses on a single dataset
and evaluate methods based on a metric,
the choice of which has varied across the years.
Most recently, the winner was chosen
based on a simple average of AUC, F1, F$0.5u$, and C@1.

\subsection{Datasets}

\citet{murauer2021developing} worked towards a benchmark
for AA,
but
overlooked the applicability of AV methods.
Moreover, they do not discuss the domain shift 
(systematic statistical differences between train and test sets)
present in many popular datasets.
Not only are these datasets not independent and identically distributed,
the test sets often contain novel topics (cross-topic - $\times_t$),
genre's (cross-genre - $\times_g$),
or authors (unique authors - $\times_a$).

Table~\ref{tab:DatasetStatistics} 
shows the wide statistical variability 
between the different datasets. 
Roughly, we can categorize the datasets into small (CCAT50, CMCC, Guardian), 
medium (IMDb62, Blogs50, Amazon), 
and large (BlogsAll, PAN20 \& 21, Gutenberg) datasets. 
The number of authors, documents, and words in a corpus is influential, 
but looking more closely at the number of documents per author ($D/A$) 
and the number of words per document ($W/D$) 
gives a better idea of how hard a corpus is. 
The larger the number of authors and 
the less text there is to work with,
the harder the problem.
Lastly, we measure the imbalance ($imb$) of datasets 
based on the standard deviation 
of the number of documents per author,
most impactful in the AA formulation. 

The cross-topic setting is typically used 
to understand the sensitivity of models to topical variations.
\citet{altakrori2021topic} observe that any errors in this setting 
can be caused by a failure to capture writing style as well as topic shift. 
They then introduce the topic-confusion setting
that can more clearly distinguish the error source. 
The topic-confusion setting is when all topics 
appear in both the training and test sets 
but the topic of the texts for each author changes in the test set
(i.e. author $A_1$ writes on topic $T_1$ 
and author $A_2$ writes on topic $T_2$
in the training set, 
then in the test set, 
author $A_1$ writes on topic $T_2$
and author $A_2$ writes on topic $T_1$).
The authors find that traditional N-gram methods
are the most robust in this setting. 
While we do not include any experimentation on topic-confused datasets,
\valla does include the topic confusion split they introduced.

The CCAT50~\cite{lewis2004rcv1}, CMCC~\cite{goldstein2008creating}, Guardian~\cite{stamatatos2013robustness},
IMDb62~\cite{seroussi2014authorship},
and PAN20 \& PAN21~\cite{kestemont2021overview} 
are used as they are in prior work, 
but with the distinction that we publish 
our train/validation/test splits 
to ensure comparability with future work. 
The Amazon review corpus~\cite{he2016ups} is a very large dataset, 
including over 142 million reviews. 
However, we extract only the users who have exactly 10 reviews.

The Blogs50 dataset is common and interesting 
due to its small average text length. 
However, the dataset was introduced 
by~\citet{schler2006effects} in its raw form, 
which correlates to the BlogsAll entry 
in Table~\ref{tab:DatasetStatistics}. 
However, the statistics we present 
are different than originally published. 
This discrepancy is due to a large number of duplicates
($\sim$160,000 exact duplicates) which we have removed. 
The most common form of this dataset is Blogs10 and Blogs50 
(the data texts from the ``top'' 10 and 50 authors respectively). 
This is problematic because it isn't clear 
how these ``top'' authors are selected: 
the number of documents~\citep{fabien2020bertaa, patchala2018authorship}, 
the number of words, with minimum text length~\citep{koppel2011authorship},
with spam (or other) filtering~\cite{yang2014authorship, halvani2017usefulness}, 
or as in most cases, not specified~\citep{Jafariakinabad2022self, yang2018topic, zhang2018syntax, ruder2016character}. 
In our framework, we release standard splits and filtering 
(only removing exact duplicates) for this dataset. 

Machine learning has benefited from larger datasets. 
However, as highlighted in Table~\ref{tab:DatasetStatistics},
there is only one common large-scale dataset: 
PAN20 \& PAN21\footnote{These datasets share a training set, 
differing only in the test set, so we treat them as a single dataset here}. 
We add a new dataset to this list: Gutenberg. 
While some prior work has used Project Gutenberg\footnote{\url{https://www.gutenberg.org}} as a dataset source (public domain books), 
they all use small subsets (\citet{arun2009stopword} use 10 authors, 
\citet{brooke2015gutentag} introduce a generic interface for the data, 
\citet{gerlach2020standardized} use the 20 most prolific authors, 
\citet{menon2011domain} use 14 authors, 
\citet{rhodes2015author} use 6 authors, 
\citet{khmelev2001using} get a 380 text subset, etc.). 
Here we have collected all single-author 
English texts from Project Gutenberg resulting 
in almost 2 billion words and a very long average document length.

\subsection{Metrics}
\label{sec:metrics}

One of the difficulties in comparing prior work 
is the use of different performance metrics. 
Some examples are accuracy~\citep{altakrori2021topic, stamatatos2018masking, Jafariakinabad2022self, fabien2020bertaa, Saedi2021Siamese, zhang2018syntax, barlas2020cross}, 
F1~\citep{murauer2021developing}, 
C@1~\cite{bagnall2015author},
recall~\cite{lagutina2021comparison},
precision~\cite{lagutina2021comparison},
macro-accuracy~\cite{bischoff2020importance},
AUC~\cite{bagnall2015author, pratanwanich2014wrote},
R@8~\citep{rivera-soto-etal-2021-learning},
and the unweighted average of F1, F0.5u, C@1, 
and AUC~\citep{manolache2021transferring, kestemont2021overview, tyo2021siamese, futrzynski2021author, peng2021encoding, bonninghoff2021o2d2, boenninghoff2020deep, embarcadero2022graph, weerasinghe2021feature}.

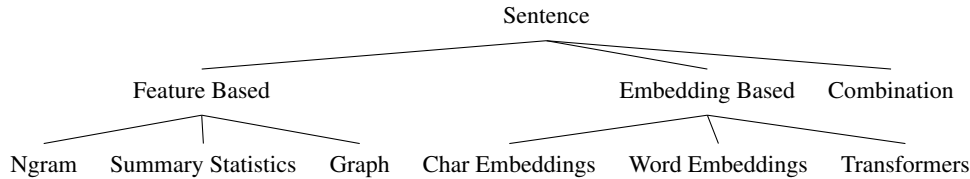
\begin{figure*}[ht]
\centering
\footnotesize
\begin{forest}
  [Sentence
    [Feature Based
     [\ngram]
     [Summary Statistics]
     [Graph]
    ]
    [Embedding Based
     [Char Embeddings]
     [Word Embeddings]
     [Transformers]
    ]
    [Combination]
  ]
\end{forest}
\caption{Hierarchy of Prior Work feature extraction methods}
\label{fig:priorTree}
\end{figure*}

The F0.5u, C@1, and Brier Score metrics were popularized by the PAN competitions 
to measure the ability of models to abstain from hard samples
in an AV problem formulation.
A participant submits a score between 0 and 1 for each sample 
with 0 indicating that the texts were written by different authors
and 1 the same author. 
To abstain from a sample, the participant 
must submit a score of exactly $0.5$. 
However, another one of the metrics used in the competition is the AUC score,
which accounts for how well a model can rank predictions
(i.e. giving a convenient measure of model performance without concern of a threshold). 
The PAN scoring ignores any abstained sampled in the AUC calculation 
instead of allowing participants to submit two numbers for each sample,
one indicating the model's score and another indicating 
if the model wishes to abstain or not 
(then the AUC can be measured without the influence of abstaining 
while the metrics that account for these non-answers 
can still be used as intended).
However, we view this level of metric specificity 
as more relevant to specific applications 
and therefore do not include them in this work. 

In both problem formulations, we want to understand
the discriminative power of each model, 
and we need to be %
careful to avoid metrics that are determined
too much by performance on a small subset
of prolific authors. 
Thus, we adopt \emph{macro-averaged accuracy} for AA (referred to as macro-accuracy), 
and \emph{AUC} for AV.

\subsection{Methods}

This section overviews prior methods organized by feature extraction method (Figure~\ref{fig:priorTree}). 

\subsubsection{Feature Based} 
\noindent \textbf{\ngram {} {}} 
The most commonly seen input representation (feature) 
used in AA and AV problems are of N-grams. 
In most cases, the N-grams are counted into a bag-of-words representation, 
but in some cases, they are transformed to feature representations using Convolutional Neural Networks (CNNs)~\citep{shrestha2017convolutional}. 
\citet{granados2011reducing} introduced \emph{text distortion},
which substitutes out-of-vocabulary items for a ``*''.
\citet{stamatatos2018masking} and \citet{bischoff2020importance} 
further test these distortion methods and more complex domain-adversarial methods, 
showing that the simpler distortion methods are most effective. 
Table~\ref{tab:textDistortion} gives examples of this text distortion.

\begin{table}[]
\begin{tabular}{@{}cc@{}}
\toprule
Given Sentences & The dog's (and cat's) house. \\ \midrule
Single Asterisk & The *'s (and cat's) *. \\
Multiple Asterisk & The ***'s (and cat's) *****. \\
Exterior Characters & The d*g's (and cat's) h***e. \\
Last Two Characters & The *og's (and cat's) ***se. \\ \bottomrule
\end{tabular}
\caption{Common text distortion methods. The vocabulary in the given example is \{The, and, cats\}.}
\label{tab:textDistortion}
\end{table}

Long considered the best AA/AV method, 
the \ngram-based \emph{unmasking} method,
developed by \citet{koppel2004authorship},
is based on the idea that the style of texts
from the same author differs only in a few features. 
Given a test text A, unmasking works 
by taking all texts from a potential author B
and then building a classifier 
to predict if the text is from A or B. 
A small number of the most meaningful features are removed, 
and this process is repeated until there are no features left to remove. 
At each step, the accuracy is tracked 
creating a performance degradation curve.
Finally, an SVM is trained to classify the degradation curve 
to determine if A and B are from the same author or not 
(the ability of the model to distinguish 
between A and B will decay quickly 
if they are from the same author,
as only a few features change between 
the works of a single author,
whereas there are many differences between the work of different authors.
\citet{koppel2011authorship} later change this method 
to keep score of how often each author is predicted after each feature elimination round, 
and then make a final prediction based on these scores, dubbed the imposter's method,
and \citet{bevendorff2019generalizing} use this method 
for short texts by oversampling each text.

\citet{seroussi2011authorship} use Latent Dirichlet Allocation (LDA),
comparing the distance between text representations to determine authorship.  
They find that this topic modeling approach 
can be competitive with the imposter's method 
while requiring less computation. 
\citet{seroussi2014authorship} expand on this topic model approach, 
and while they
presents good results on the PAN'11 dataset, 
the performance of the topic modeling approaches lags 
behind the best methods.

\citet{zhang2018syntax} introduce a high-performing method 
that leverages sentence syntax trees and
character n-grams as input to a CNN. 
\citet{Saedi2021Siamese} also presents good results with CNN models,
but \citet{ordonez2020will} indicate 
that these CNN methods are no longer competitive.

\noindent \textbf{Summary Statistics {} {}} 
While older methods focused on small sets of summary statistics, 
more modern methods are able to combine all of these into a single model.
\citet{weerasinghe2021feature} provide the best example of this,
calculating a plethora of hand-crafted features and Ngrams for each document
(distribution of word lengths, hapax-legomena, Maas’ a$^2$, Herdan's V$_m$, and more). 
The authors take the difference between these large feature vectors for two texts 
and then train a logistic regression classifier 
to predict if the texts were written by the same author or not. 
Despite its simplicity, this method performs well.

\noindent \textbf{Co-occurance Graphs {} {}} 
\citet{arun2009stopword} construct a graph 
that represents a text based purely on the stopwords (nodes) 
and the distance between them (edge weights). 
Then to compare the two texts, their graphs 
are compared using the Kullback-Leibler (KL) divergence. 
\citet{embarcadero2022graph} also construct a graph for each text i8
but instead represents each node as a [word, POS\_tag] tuple,
and each vertex indicates adjacency frequency. 
After the graph is created for each text it is encoded 
into a one-hot representation and used as input to a LEConv layer.
After pooling, the absolute difference between the two document representations
is passed through a five-layer fully connected network for final scoring.

\subsubsection{Embedding Based}
\noindent \textbf{Char Embedding {} {}} \citet{bagnall2015author} use 
a character-level recurrent neural network (RNN) for authorship verification 
by sharing the RNN model across all authors but training a different head 
for each author in the dataset.
To classify authors, they calculate the probability 
that each text was written by each author, 
predicting the author with the highest probability. 
Interestingly, the authors accidentally 
ran a version of their system without the RNN (just the multi-heads) 
and achieved competitive results. 

\citet{ruder2016character} use both CNNs 
to embed both characters and words for AA. 
Their results show that the character-based method 
outperforms the word-based approach across several datasets. 
Compression-based methods, which leverage a compression algorithm 
(such as ZIP, RAR, PPM, etc.) to build text representations 
which are then compared with a distance metric,
fall into this category as well~\citep{halvani2017usefulness}.

\noindent \textbf{Word Embedding {} {}}
\citet{boenninghoff2019explainable} leverage 
the Fasttext pre-trained word embeddings, 
concatenated with a learned CNN character embedding,
as part of the input to a hierarchical bi-directional 
Long Short Term Memory (BiLSTM) network. 
Specifically, this combined embedding is used as input 
to a word-to-sentence BiLSTM network, 
whose output is fed 
into a sentence-to-document BiLSTM
to produce a final document embedding. 
This neural network structure runs in parallel for two documents 
(i.e. as a Siamese network~\cite{koch2015siamese}),
and then optimized according to the modified contrastive loss function 
(i.e. the traditional contrastive loss 
but additionally doesn't penalize same-author samples 
that are sufficiently close in the embedding space).
This method was introduced by \citet{boenninghoff2019explainable}, 
and then later modified to include Bayes factor scoring 
on the output by \citet{boenninghoff2020deep}, 
and by \citet{bonninghoff2021o2d2}
to include an uncertainty adaptation layer for defining non-responses. 
This was the highest performing method at the PAN20 and PAN21 competitions~\citep{kestemont2021overview}.

\citet{Jafariakinabad2022self} attempted to build the equivalent 
of pre-trained word embeddings but for sentence structure 
(i.e. GloVe-like embeddings that map sentences with a similar structure
close together but are agnostic of their ``meaning'').
They learn these embeddings by creating a parse-tree 
for each sentence using the CoreNLP parser 
and then passing this parse-tree 
and a traditional word-embedded sentence 
through identical but separate BiLSTMs 
and train via contrastive loss.
Then, for authorship attribution, 
they pass the sentence through 
both a typical word-embedding LSTM, 
and also through their learned sentence structure encoding network, 
combining their output. 
The authors also compare against their prior work~\citep{jafariakinabad2019style} 
which embeds the POS-tags along with the word embeddings
instead of using their custom structural embedding network.
The new work slightly outperforms their prior method 
and is more efficient as you can leverage 
the pretrained structure encoding network 
instead of having to always label the POS tags.

CNN's have also been well explored given word embeddings as 
input~\citep{hitschler2018authorship, shrestha2017convolutional, ruder2016character}, 
yet their results are not among the highest reported.

\noindent \textbf{Transformers {} {}} \citet{rivera-soto-etal-2021-learning} attempt to build 
universal representations for AA and AV 
by exploring the zero-shot transferability 
of different methods between three different datasets. 
The authors train a Siamese BERT model~\citep{reimers2019sentence} on one dataset 
and then test the performance on another dataset 
without updating the model on this new dataset. 
Unfortunately, the results seem 
to indicate more about the underlying datasets 
then the ability of these models 
to uncover a universal authorship representation. 

\citet{manolache2021transferring} also explore
the applicability of BERT to AA 
by using BERT embeddings as the feature set 
for the aforementioned unmasking method. 
Comparing this to Siamese BERT, 
Character BERT~\citep{boukkouri2020characterbert}, 
and BERT for classification, 
they find that simple fine-tuning 
outperforms the more complicated unmasking setup.

Following \citet{bagnall2015author}, \citet{barlas2020cross} 
approach the AA problem by using a shared language model 
with a different network head for each author. 
They then compare different shared language model architectures 
(RNN, BERT, GPT2, ULMFiT, and ELMo), 
finding that pretrained language models
improve the performance of the original RNN architecture. 
However, the results are all from the small CMCC corpus.

\citet{tyo2021siamese} use a Siamese BERT setup 
with triplet loss and hard-negative mining for training. 
\citet{futrzynski2021author} concatenate 28 tokens from each text
and then use BERT's \texttt{[CLS]} output token for author classification.
\citet{peng2021encoding} concatenate 256 tokens from each text 
to produce a 512 token input for BERT, 
and then after pooling use linear layers for same/different author prediction.
They repeat this 30 times, sampling different sections of the input texts, 
and then average over the 30 predictions for final classification. 

\subsubsection{Feature and Embedding Based}

\citet{fabien2020bertaa} were an early work 
exploring the applicability of BERT to authorship attribution. 
They combine the output of BERT with summary statics 
via a logistic regression classifier. 
The authors find that a BERT-only model was as effective 
as a model combining the BERT output with the summary statistics.

\section{The \valla Benchmark}
\label{sec:valla}

In 1440, Lorenzo Valla
proved that the
\textit{Donation of Constantine}
(where Constantine \RNum{1} gave the whole of the Western Roman Empire to the Roman Catholic Church) was a forgery, using word choice and other vernacular stylistic choices as evidence~\citep{valla1922discourse}. 
Inspired by this influential use of AA, we introduce \valla: A standardized benchmark for authorship attribution and verification.\footnote{Valla can be found here: \url{https://github.com/JacobTyo/Valla}}
\valla includes all datasets in Table~\ref{tab:DatasetStatistics},
along with others from prior literature
\citep{klimt2004enron, manolache2022veridark, overdorf2016blogs, altakrori2021topic},
with standardized splits, cross-topic/cross-genre/unique author
test sets,
and usable in either AA or AV formulation.
\valla also includes five method implementations, and we use the subscript ``A'' or ``V'' to distinguish between the attribution and verification model formulations respectively.

\noindent \textbf{\ngram {} {}} Being the best performing method in \citet{altakrori2021topic}, 
\citet{murauer2021developing}, \citet{bischoff2020importance}, and \citet{stamatatos2018masking}, 
this method creates character \ngram, part-of-speech \ngram, 
and summary statistics for use as input to an ensemble of logistic regression classifiers. 
For use in the AV setting, we follow \citet{weerasinghe2021feature} by using the difference between the \ngram~feature vectors of two texts as input to the logistic regression classifier. 

\noindent \textbf{PPM {} {}} Originally developed in \citet{Teahan2003Compression} 
and best performing in \citet{neal2017surveying}, 
this method uses the prediction by partial matching (PPM) compression model (a variant of PPM is used in the RAR compression software) to compute a character-based language model for each author \citep{halvani2018cross},
and then the cross-entropy between a test text and each author model is calculated.
For use in an AV setup, one text is used to create a model and then the cross-entropy is calculated on the second text. 

\noindent \textbf{BERT {} {}} With the highest reported performance on the AA dataset Blogs50~\citep{fabien2020bertaa}, 
this method combines a BERT pre-trained language model with a dense layer for classification. 
For evaluation, we chunk the evaluation text into non-overlapping sets of 512 tokens and take the majority vote of the predictions. 
For use in the AV setup, the BERT model is used as the base for a Siamese network and trained with contrastive loss. 
For evaluation in the AV setup, we chunk the two texts into $K$ sets of 512 stratified tokens (such that the first 512 tokens of each text are compared, the second grouping of 512 tokens from each text is compared, etc.), and then take the majority vote of the $K$ predictions. 

\noindent \textbf{pALM {} {}} The best performing model in \citet{barlas2020cross} was another variation on BERT where a different head was learned on top of the BERT language model for each known author. 
We refer to this method as the per-Author Language Model (pALM). 
To classify a text, it is passed through the model for each author, and then the author model with the lowest perplexity on the text is predicted.
This is only used in AA formulations as in AV we would have only a single text to train a network head with.

\noindent \textbf{HLSTM {} {}} Originally introduced by \citet{boenninghoff2019explainable},  this method leverages a hierarchical BiLSTM setup with Fasttext word embeddings along with custom word embeddings learned using a character level CNN as the base for a Siamese network.
This method was the highest performing at PAN20 and PAN21~\cite{kestemont2021overview} and is only used in AV formulations. 
While this can be modified to work in AA, we follow prior work and use the original AV setup only.

All of these methods fall into two categories: 
the methods that predict an author class, 
and the methods that predict text similarity. 
The methods that predict an author class (whether via logistic regression, dense layer, SVM, etc.) need no post-processing. 
However, the methods that predict similarity need post-processing both for AA and AV problems. 
For AA, we build an \emph{author profile} by randomly selecting 10 texts from each author and averaging their embeddings together. 
Then we can compare the unknown texts to each author profile and predict the author that is most similar (using the euclidean distance).
For AV, we directly compare the text representations (again using euclidean distance) and then define a hard threshold based on a grid search on the evaluation set
(although for computing AUC this threshold is irrelevant).

\begin{table*}[t]
\centering
\begin{tabular}{@{}ccccccccc@{}}
\toprule
           & CCAT50 & CMCC  & Guardian & IMDb62 & Blogs50 & PAN20 & Gutenburg AA & Average \\ \midrule
\ngrama & 76.68  & 86.51 & 100      & 98.81  & 72.28   & 43.52 & 57.69        & 76.50   \\
PPM\aam    & 69.36  & 62.30 & 86.28    & 95.90  & 72.16   & ---   & ---          & 55.14   \\
BERT\aam   & 65.72  & 60.32 & 84.23    & 98.80  & 74.95   & 23.83 & 59.11        & 66.71   \\
pALM\aam   & 63.36  & 54.76 & 66.67    & ---    & ---     & ---   & ---          & 26.40   \\ \bottomrule
\end{tabular}
\caption{Macro-accuracy (\%) of the authorship attribution models. The ``Average''column represents the average macro-accuracy of each model across all datasets in this table, where --- entries are counted as 0\%.
}
\label{tab:aaPerformance}
\end{table*}

\begin{table}[tb]
\centering
\begin{tabular}{ccccc}
\hline
 &
  \begin{tabular}[c]{@{}c@{}}CMCC\\ $\times_t$\end{tabular} &
  \begin{tabular}[c]{@{}c@{}}CMCC\\ $\times_g$\end{tabular} &
  \begin{tabular}[c]{@{}c@{}}Guard\\ $\times_t$\end{tabular} &
  \begin{tabular}[c]{@{}c@{}}Guard\\ $\times_g$\end{tabular} 
   \\ \hline
\ngrama & 82.54 & 84.13 & 86.92 & 87.22 \\
PPM\aam                                                       & 52.38 & 57.14 & 69.23 & 72.08 \\
BERT\aam                                                    & 49.21 & 45.24 & 75.64 & 75.56 \\
pALM\aam                                                  & 57.14 & 46.03 & 61.79 & 47.22 \\ \hline
\end{tabular}
\caption{Macro-accuracy (\%) of the authorship attribution models.}
\label{tab:aaPerformanceCross}
\end{table}

\begin{table}[bt]
\centering
\begin{tabular}{@{}ccccc@{}}
\toprule
                   & PAN21  & AmaAV & BlogAV & GutAV  \\ \midrule
\ngramv & 0.9719 & 0.7742 &  0.5410 & 0.8741 \\
PPM\avm                & 0.7917 & 0.6492 &  0.6230 & 0.8508 \\
BERT\avm       & 0.9709 & 0.8943 &  0.9201 & 0.9624 \\
HLSTM\avm          & 0.9693 & 0.8734 &  0.8580 & 0.9147 \\ \bottomrule
\end{tabular}
\caption{AUC of the AV models on the selected AV datasets.}
\label{tab:avPerf}
\end{table}

\begin{table*}[t]
\centering
\begin{tabular}{@{}cccccc@{}}
\toprule
 & CCAT50 & CMCC & Guardian & IMDb62 & Blogs50 \\ \midrule
HLSTM\avm & 4.56 & 8.33 & 27.59 & 37.82 & 57.49 \\
(P)HLSTM\avm & 13.36 & 16.27 & 38.97 & 59.47 & 11.34 \\
BERT\avm & 48.64 & 35.75 & 27.82 & 76.62 & 60.72 \\
(P)BERT\avm & 56.80 & 40.87 & 61.41 & 73.17 & 67.21 \\ \midrule
BERT\aam & 65.72 & 60.32 & 84.23 & 98.80 & 74.95 \\ \bottomrule
\end{tabular}
\caption{Macro-accuracy (\%) of the AV models on AA datasets. The (P) indicates that the model was pretrained on the PAN20 training set before fine-tuned on the corresponding dataset.
}
\label{tab:avOnAaPerformance}
\end{table*}

\begin{table}[tb]
\centering
\begin{tabular}{@{}ccccc@{}}
\toprule
 & \begin{tabular}[c]{@{}c@{}}CMCC\\ $\times_t$\end{tabular} & \begin{tabular}[c]{@{}c@{}}CMCC\\ $\times_g$\end{tabular} & \begin{tabular}[c]{@{}c@{}}Guard\\ $\times_t$\end{tabular} & \begin{tabular}[c]{@{}c@{}}Guard\\ $\times_g$\end{tabular} \\ \midrule
HLSTM\avm & 7.94 & 3.18 & 19.23 & 23.33 \\
(P)HLSTM\avm & 9.52 & 5.56 & 40.00 & 31.53 \\
BERT\avm & 28.85 & 13.49 & 42.31 & 46.53 \\
(P)BERT\avm & 33.33 & 19.05 & 43.33 & 54.72 \\ \midrule
BERT\aam & 49.21 & 45.24 & 75.64 & 75.56 \\ \bottomrule
\end{tabular}
\caption{Macro-accuracy (\%) of the authorship verification models on the domain shift AA datasets. The (P) indicates that the model was pretrained on the PAN20 training set before fine-tuned on the corresponding dataset.
}
\label{tab:avOnaaCrossData}
\end{table}

\begin{table}[tb]
\centering
\begin{tabular}{@{}cccc@{}}
\toprule
\begin{tabular}[c]{@{}c@{}}Metric\\ (Formulation)\end{tabular} & \begin{tabular}[c]{@{}c@{}}AUC\\ (AV)\end{tabular} & \begin{tabular}[c]{@{}c@{}}Acc\\ (AV)\end{tabular} & \begin{tabular}[c]{@{}c@{}}Mac-Acc\\ (AA)\end{tabular} \\ \midrule
BERT\avm & 0.9229 & 82.33 & 67.21 \\
BERT\avm w/HNM & 0.9276 & 82.72 & 72.42 \\ \bottomrule
\end{tabular}
\caption{This table compares the performance of the same model (BERT\avm), on the same data (Blogs50), just formulated in different ways, using different performance metrics (column header). w/HNM represents training with hard negative mining.
}
\label{tab:AaAVCompare}
\end{table}

\section{Experiments and Discussion}
\label{sec:results}

\subsection{The State-of-the-Art in Authorship Attribution}

After evaluating 
all methods in \valla on the AA datasets listed in Table~\ref{tab:DatasetStatistics}, 
we find that the traditional \ngram ~method is the highest performing on average as detailed in Table~\ref{tab:aaPerformance}. 
However, we do see that the BERT\aam model closes the gap on (and can even exceed) 
the performance of the \ngrama method as the size of the training set increases. 
This correlation does not hold on the PAN20 dataset, 
where the best performing model is still \ngrama.
This indicates that the state-of-the-art AA method is dependent upon the number of words per author available. 
While we do not provide a detailed analysis of the data requirements of each method, our results roughly indicate that \ngrama is the method of choice for datasets with less than $50,000$ words per author, while BERT\aam is the state-of-the-art method for datasets with over $100,000$ words per author. 

PPM\aam is simple to tune due to few hyperparameters, 
but it is both a low performer and it scales poorly to large datasets (rendering it unusable on the PAN20 and Gutenberg datasets). 
pALM\aam is the lowest performing method tested, is expensive to train, and scales poorly resulting in too slow of training to be used on the IMDb62, Blogs50, PAN20, and Gutenberg datasets.

The macro-accuracy of BERT\aam on the IMDb62 and Blogs50 datasets presents a new state of the art,
while defining the initial performance marks on the GutenbergAA and PAN20 datasets.\footnote{These are initial results because the PAN20 competition was formulated as an AV problem, whereas here we use the AA formulation}
The performance on the Blogs50 dataset requires a bit more analysis due to our filtering of duplicates in the dataset. 
As a better comparison to prior reported performance, we first explore the performance of BERT\aam
on the Blogs50 dataset \emph{without} the filtering, and achieve a macro-accuracy of $64.3\%$. 
This represents the state-of-the-art accuracy on a version of the dataset more comparable with prior work (despite its issues) but indicates the strength of the result reported in Table~\ref{tab:aaPerformance}. 

Our results on the Guardian and CMCC datasets are hard to compare to prior work due to the previously mentioned standardization issues, most notably a i.i.d. split has not been used in prior work. 
The CCAT50 dataset, on the other hand, is directly comparable to prior work. 
Currently, we show best performing model as the \ngram. 
However, \citet{Jafariakinabad2022self} reports the accuracy of a CNN that takes the syntactic tree of a sentence as input as $83.2\%$ which is better than what we were able to achieve.\footnote{CCAT50 is a balanced dataset, so the macro-accuracy and accuracy are equal.} 
Therefore we report this previous best result on the performance tracking page of \valla.

\subsection{The State-of-the-Art in Authorship Attribution under Domain Shift}

While the problem of achieving high performance under any domain shift is an open problem in machine learning, 
exploration of domain shift in AA and AV settings is common even if not always explicitly recognized. 
Table~\ref{tab:aaPerformanceCross} examines the performance of the same AA models discussed in the previous section but focuses on the cross-topic and cross-genre test sets of the CMCC and Guardian datasets.
Just as in the i.i.d. setting, the \ngrama method dominates in all scenarios. 
It should be noted that all datasets used in this domain shift scenario are small, 
so we cannot verify that the BERT\aam method would begin to dominate as the number of words per author increases.  %
We leave the exploration of domain shift performance on larger datasets to future work, 
although we expect that the BERT\aam model would begin to outperform \ngrama.

\subsection{The State-of-the-Art in Authorship Verification}

In keeping in line with prior work, the distinction between datasets with and without domain shift is less clear when formulated as an AV problem. 
The PAN21 test set is comprised of authors that do not appear in the training set. 
However the remainder of the datasets (AmazonAV, BlogsAV, and GutenbergAV) are all iid dataset splits. 
Table~\ref{tab:avPerf} Details the performance of the AV methods on selected AV datasets. 

While we saw the \ngrama method dominating on most AA datasets, here we see that the deep learning-based HLSTM\avm and BERT\avm methods attain the highest AUC across the board.
However, as we saw the bigger models outperform smaller models when the number of words per class became large, 
in AV there are only two classes (same and different author), and therefore all of the datasets have a very large number of words per class. 
Because of this key difference, AV formulations tend to be more effective for training deep learning methods.

\subsection{Comparing AA and AV methods}

The AA formulation is common in practice, and many AV works are motivated by the idea that they can be applied to these common AA problems. 
Yet despite the prominence of comments indicating how AV is the fundamental problem of AA 
(seemingly hinting that applying AV methods to AA problems is easy), 
there is no evidence of how well their performance actually transfers. 
We dive into this situation in this section. 
Table~\ref{tab:avOnAaPerformance} shows the performance of LSTM\avm and BERT\avm on the i.i.d. AA datasets, both when trained only on the dataset as well as starting from a pretrained version of the models (the PAN20 training set was used for pretraining). 
Here we see notably lower performance than what was obtained by the AA methods,
proving that the application of AV methods to AA problems is not as straightforward as it may seem. 
Table~\ref{tab:avOnaaCrossData} shows this trend continuing in the cross-topic and cross-genre AA evaluation settings as well.\footnote{We note that the lower performance of the pretrained H-LSTM on Blogs50 than its non-pretrained version is due to the vocabulary selection. 
This method chooses its vocabulary based on the pretraining corpus, 
and therefore runs the risk of the vocabulary not transferring well to the intended corpus.}

\subsection{Hard-Negative Mining}

As indicated in the previous section, 
although AV is the fundamental problem of AA, 
AV methods do not necessarily perform well under an AA formulation.
To correctly classify a text in the AA setting, 
a model must make harder comparisons 
(i.e., it must compare a text to the hardest negative---the 
text that makes the decision the hardest),
whereas an AV setting is strictly easier 
as it must compare to only a single text. 
AV problems can be made harder, 
but they cannot ever consist of exactly the hardest negatives, 
because the hardest negative is model-dependent. 
This interpretation motivates the exploration of using hard-negative mining
(updating a model during training only on the hardest examples in each batch)
for improving the transferability of AV methods to AA problems. 

The AV formulation is very attractive for training deep models, so bridging the performance gap would create a strong case for more research emphasis on AV models. 
In this section we take a single model (BERT\avm) and train two versions of it: one with the contrastive loss and one using triplet loss with batch hard negative mining (specifically the per-batch hard negative mining methodology used in \citet{hermans2017defense}).
Then using our standardized Blogs50 dataset, we can determine the AV AUC to AA macro-accuracy relation. 

Table~\ref{tab:AaAVCompare} details these results, showing two key findings. 
The first is that high AV AUC does not indicate high AA macro-accuracy, 
and the second is that training an AV method with hard negative mining has little effect on its AV AUC but drastically improves its AA macro-accuracy.

\section{Conclusion}
\label{sec:conclusion}

After a survey of the AA and AV landscapes, 
we present \valla: an open-source dataset and metric standardization benchmark, complete with method implementations. 
Using \valla and a newly introduced large-scale dataset, we present an extensive evaluation of AA and AV methods in a wide variety of common formulations.
We achieve a new state-of-the-art macro-accuracy on the IMDb62 ($98.81\%$) and Blogs50 ($74.95\%$) datasets and provide benchmark results on the other datasets. 
Among sufficient words per author ($\sim 100,000$), the state-of-the-art AA method is BERT\aam, 
but if ample data is not available the \ngrama method proves most effective. 
When faced with domain shift, we find the \ngrama method to be the most accurate,
but notably, all of the domain-shifted datasets are small. 

Our results show that the AV problem formulation is more effective for training deep models, finding the state-of-the-art AV method to be BERT\avm. 
After showing that the high-performing BERT\avm does not perform competitively in AA problems, 
we explore the effect of hard-negative mining on its performance and find that with no degradation in AV performance, it improves the AA macro-accuracy of BERT\avm by over $5\%$, making it a competitive method in the AA formulation. 
We hope that \valla encourages more work on understanding the AA and AV method landscapes, developing new AA/AV methods, and allows for direct comparison of future findings.

\clearpage

\bibliography{references}
\bibliographystyle{acl_natbib}

\end{document}